\documentclass{article}

\usepackage{microtype}
\usepackage{graphicx}
\usepackage{subcaption}
\usepackage{booktabs} % for professional tables

\usepackage{hyperref}
\usepackage{multirow}

\usepackage[preprint]{icml2026}

\usepackage{amsmath}
\usepackage{amssymb}
\usepackage{mathtools}
\usepackage{amsthm}

\usepackage[capitalize,noabbrev]{cleveref}

\theoremstyle{plain}

\theoremstyle{definition}

\theoremstyle{remark}

\usepackage[textsize=tiny]{todonotes}

\icmltitlerunning{Deep Microcompression}

\begin{document}

\twocolumn[
  \icmltitle{Deep Microcompression: Structured Pruning and Bit-packed Quantization for Microcontrollers}

  % It is OKAY to include author information, even for blind submissions: the
  % style file will automatically remove it for you unless you've provided
  % the [accepted] option to the icml2026 package.

  % List of affiliations: The first argument should be a (short) identifier you
  % will use later to specify author affiliations Academic affiliations
  % should list Department, University, City, Region, Country Industry
  % affiliations should list Company, City, Region, Country

  % You can specify symbols, otherwise they are numbered in order. Ideally, you
  % should not use this facility. Affiliations will be numbered in order of
  % appearance and this is the preferred way.
  \icmlsetsymbol{equal}{*}

  \begin{icmlauthorlist}
    \icmlauthor{Opegbemi Matthias Busoye}{comp}
    \icmlauthor{Tolulope Matthew Busoye}{comp}
    \icmlauthor{Eghonghon-aye Eigbe}{comp}
  \end{icmlauthorlist}

  % \icmlaffiliation{yyy}{Department of XXX, University of YYY, Location, Country}
  \icmlaffiliation{comp}{PowerLabs Technologies, Lagos, Nigeria}
  % \icmlaffiliation{sch}{School of ZZZ, Institute of WWW, Location, Country}

  \icmlcorrespondingauthor{Opegbemi Matthias Busoye}{matthias@powerlabstech.com}
  \icmlcorrespondingauthor{Tolulope Matthew Busoye}{matthew@powerlabstech.com}

  % You may provide any keywords that you find helpful for describing your
  % paper; these are used to populate the "keywords" metadata in the PDF but
  % will not be shown in the document
  \icmlkeywords{TinyML, Model Compression, Bare-Metal Inference, Structured Pruning, Quantization, Bit-Packing}

  \vskip 0.3in
]

% this must go after the closing bracket ] following \twocolumn[ ...

% This command actually creates the footnote in the first column listing the
% affiliations and the copyright notice. The command takes one argument, which
% is text to display at the start of the footnote. The \icmlEqualContribution
% command is standard text for equal contribution. Remove it (just {}) if you
% do not need this facility.

% Use ONE of the following lines. DO NOT remove the command.
% If you have no special notice, KEEP empty braces:
\printAffiliationsAndNotice{}  % no special notice (required even if empty)
% Or, if applicable, use the standard equal contribution text:
% \printAffiliationsAndNotice{\icmlEqualContribution}

\begin{abstract}
This paper introduces Deep Microcompression (DMC), a hardware-aware pipeline for deep learning inference on bare-metal microcontrollers. DMC integrates structured pruning, quantization-aware training, and fixed-length bit-packing to achieve a 55.8$\times$ weight compression ratio on LeNet-5 (98.77\% accuracy), generating a dependency-free C library with deterministic latency. On the RP2040 (Cortex-M0+), DMC reduces binary size by 3$\times$ versus TensorFlow Lite while matching its accuracy. Critically, DMC enables the first documented deployment of a standard CNN on the ATmega328P, a device constrained to 2KB SRAM, previously considered infeasible for CNN inference.
\end{abstract}

\section{Introduction}

The rapid evolution of Artificial Intelligence (AI) over the last decade is increasingly pushing computational workloads from the cloud to the edge, enabling local, energy-efficient intelligence in a new generation of embedded systems. This paradigm, known as TinyML, focuses on deploying machine learning models on ultra-low-power microcontrollers (MCUs). However, this trend has exposed a fundamental challenge: a significant and growing gap exists between the resource requirements of modern neural networks and the severe constraints of the hardware designed to run them. 

MCUs are defined by their strict memory, power, and compute limitations. For instance, popular platforms such as the ATmega328P feature just 2 KB of RAM and 32 KB of flash memory, while the more powerful RP2040 offers 264 KB of RAM and 2 MB of flash. Deploying full-scale, unoptimized models on these bare-metal systems is simply infeasible. To address this, various compression strategies have been developed, including pruning, quantization, binarization and knowledge distillation, which aim to reduce a model's size and operational complexity \cite{zhang2023review}. 

The influential Deep Compression framework achieved significant size reduction through a combination of pruning, trained quantization, and Huffman coding \cite{han2015deep}. However, its methods relied on complex, variable-length coding and unstructured pruning, which require specialized software decoders or hardware accelerators due to, (i) irregular memory access patterns and, (ii) additional control overhead introduced because standard dense matrix multiplication kernels on general-purpose CPUs or microcontrollers \cite{9302790, gale2019statesparsitydeepneural, 10.1145/3613424.3614303} cannot efficiently process the decoding. These requirements critically limit their deployability for real-time inference on bare-metal microcontrollers. This limitation highlights a significant gap: the need for a unified compression pipeline explicitly designed to reconcile the theoretical efficiency of deep learning compression methods with the rigid architectural constraints of bare-metal devices.

This paper introduces the Deep Microcompression (DMC) pipeline to bridge this gap, proposing a method that is simple, effective, and directly addresses the unique challenges of bare-metal development. Our approach is a cohesive end-to-end pipeline that integrates structured pruning, quantization-aware training, and a practical hardware-aware bit-packing scheme. This pipeline is specifically designed to be framework-independent, generating a minimal C-based model that is more portable than solutions relying on large runtime libraries. The bit-packing method serves as an efficient alternative to complex techniques like Huffman coding by using simple bitwise operations for decompression on any standard MCU.  

\paragraph{Relation to Prior Work.}
The seminal Deep Compression framework~\cite{han2015deep} demonstrated that pruning, quantization, and Huffman coding can achieve large compression ratios, but its variable-length encoding introduces non-deterministic latency incompatible with bare-metal MCU. TF Lite Micro~\cite{david2021tensorflowlitemicroembedded} provides a runtime for embedded inference but incurs significant binary overhead ($>$250KB), exceeding the flash budget of 8-bit devices. DMC addresses both limitations through fixed-length bit-packing and dependency-free code generation.

\paragraph{Global South Motivation.}
In much of the Global South, legacy 8-bit MCUs like the 
ATmega328P are already embedded in low-cost agricultural, 
medical, and IoT deployments --- not the Raspberry Pi Zero, 
which demands Linux, an SD card, and higher sustained power 
draw unsuitable for battery-powered field use. By enabling 
CNN inference on \$2 hardware that practitioners already 
own, DMC eliminates cloud dependency entirely, allowing a 
student or engineer in Lagos or Nairobi to deploy a trained 
model with no internet connection and no specialized 
accelerator required.

\section{The Deep Microcompression (DMC) Method}

% First image: full-width figure*
\begin{figure*}[htb]
    \centering
    \includegraphics[width=.87\textwidth]{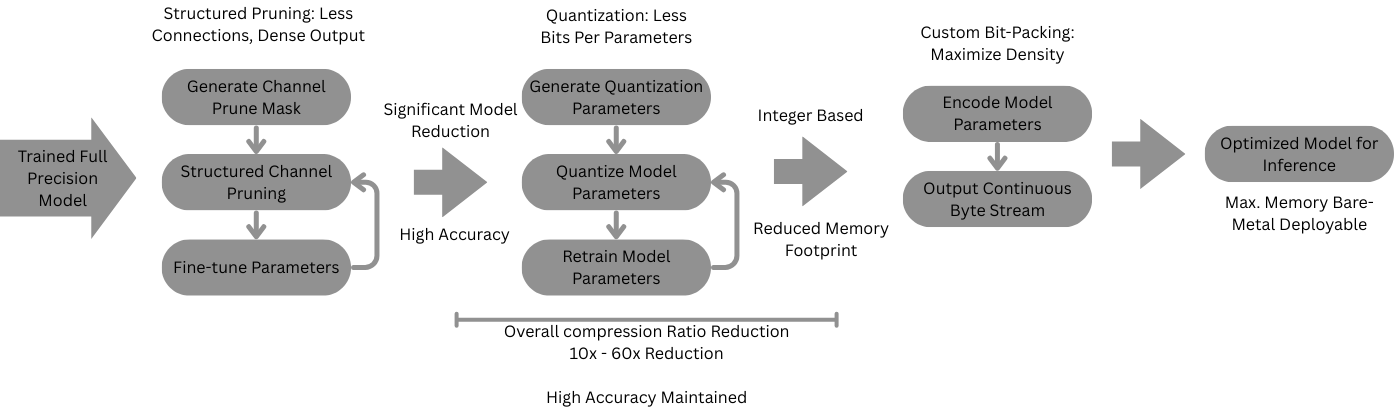}
    \caption{Deep microcompression: Development Pipeline}
    \label{fig:development-pipeline}
\end{figure*}

% Second image: single-column figure with text beside it
\begin{figure}[tb!]
    \begin{minipage}[b]{1\columnwidth} % image column
        \centering
        \includegraphics[width=\textwidth]{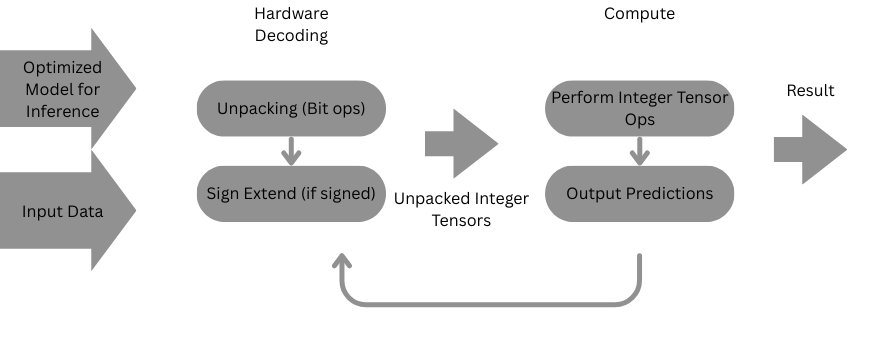}
        \caption{Deep microcompression: Inference Pipeline}
        \label{fig:deployment-pipeline}
    \end{minipage}%
\end{figure}

The DMC pipeline is built on a microcontroller-centric design philosophy with the explicit goal of enabling deep learning on bare-metal systems without requiring expensive, specialized hardware. 
This means avoiding computationally expensive tasks, complex data structures, unnecessary memory movement, and the need for specialized hardware or software decoders.

The DMC framework operates in two phases: model development (Figure~\ref{fig:development-pipeline}) and on-device inference (Figure~\ref{fig:deployment-pipeline}). The development pipeline first transforms a full-precision model through structured pruning to ensure hardware efficiency, followed by quantization-aware training to reduce parameter bitwidth, and finally utilizes a custom bit-packing scheme to consolidate multiple weights into single bytes for maximum storage density. At runtime, the inference engine decodes these packed parameters and encodes activations using low-cost bitwise operations to execute the forward pass entirely via integer operations. To enforce the integer based operations, we utilize computationally efficient activations like ReLU while complex nonlinearities like Softmax are approximated using precomputed Look-up Tables (LUTs) stored in Flash.
% A detailed explanation of each stage follows.

\subsection{Structured Pruning Stage}
To maintain architectural compatibility with standard dense matrix kernels, the initial stage of the DMC pipeline employs structured channel pruning for parameter reduction. Following the methodology in \cite{li2017pruningfiltersefficientconvnets}, we utilize an L2-norm magnitude criterion to identify and remove the least important filters in convolutional layers and neurons in fully connected layers followed by retraining for performance recovery. This ensures that the pruned model retains a dense, contiguous memory layout, preserving deterministic execution latency without requiring specialized sparse linear algebra libraries.

\subsection{Quantization Stage}
\label{sec:quantDMC}
Quantization in the DMC pipeline is designed to favor integer-based model representations over floating-point due to their superior performance on low-power devices in terms of time and energy consumption. We employ Quantization-Aware Training (QAT) to simulate discretization noise in the training phase, allowing the network to adapt to low-precision constraints. The method utilizes static quantization, a deliberate choice over dynamic quantization for both weights and activations. Static quantization pre-computes all scaling factors during a calibration phase, effectively "freezing" the dynamic range enabling a fully integer-based inference pipeline. This approach avoids the runtime overhead associated with dynamic quantization, where parameters are computed on the fly, eliminating floating-point operations from the inference stage.

\subsection{The Hardware-Aware Low-Level Optimization} \label{sec:bit_packing_method} 
A key component of the DMC pipeline is the automated generation of bit-packed inference kernels for weights and activations. While storing multiple weights per byte is a known concept, existing implementations often rely on runtime calculation of bit-offsets or heavy decoder libraries. DMC differentiates itself by shifting this complexity to the compilation stage.

The pipeline generates a dependency-free C header where the layout of every tensor is pre-calculated, serving as a practical alternative to complex schemes like Huffman coding \cite{han2015deep} or bit-serial processing \cite{li2022bitserialweightpoolscompression}. By enforcing a fixed-length storage format (e.g., $4\times$2-bit weights per \texttt{uint8}), we maximize memory density while ensuring that every decompression operation executes in constant time (O(1)) independent of bitwidth, a strict requirement for real-time bare-metal inference.

\subsubsection{Bit-Packing Strategy} To maximize storage density on resource-constrained microcontrollers, we implement a deterministic packing routine. The compiler maps a list of quantized integers into contiguous standard memory units. This structure eliminates the need for look-up tables (LUTs) for address decoding or variable length decoding. The procedure ensures that the endianness of the packed byte matches the target MCU's architecture, preventing runtime byte-swapping overhead. The specific procedures for packing individual quantized scalars and flattened tensors are detailed in Algorithm~\ref{algorithm:pack params into byte}.

\subsubsection{Optimized Bit-Unpacking} \label{sec:bit_unpacking} The unpacking stage retrieves parameter values during inference. A naive implementation would calculate the bit-offset for every weight at runtime using division and modulo operations, which are computationally expensive on 8-bit and 16-bit MCUs. To ensure minimal latency, DMC replaces these arithmetic operations with compile-time computed constants. The extraction logic is baked into the generated C code using efficient bitwise shifts and masks.
\paragraph{Unsigned Integer Datatype} For unsigned values, decompression is a single-step shift-and-mask operation. As detailed in Algorithm~\ref{algorithm:bit_unpack_method}, the packed byte is retrieved, and the target bits are isolated using pre-calculated offsets.

For example, determining the bit-offset within a byte is optimized as follows:

\begin{center}
    \small
        \begin{tabular}{lll}
        \toprule
        \textbf{Operation} & \textbf{4-bit} & \textbf{2-bit} \\
        \midrule
        $i \div n_b$ & $i >> 1$ & $i >> 2$ \\
        $b_{offset} \gets (i \bmod n_b) \times b$ & $i ~\&~ \texttt{0b1} << 2$ & $i ~\&~ \texttt{0b11} << 1$ \\
        \bottomrule
        \end{tabular}
\end{center}

\paragraph{Signed Integers:} Standard signed integers (e.g., int8\_t) rely on the Most Significant Bit (MSB) to indicate polarity so signed integers require an additional step to handle the sign bit correctly in a larger memory space (e.g., representing a 4-bit negative number in an 8-bit container). For signed data, step 6 of Algorithm~\ref{algorithm:bit_unpack_method} is modified to pass the extracted raw bits into a sign-extension routine.

\section{Experimental Results}

\begin{table*}[htbp!]
    \centering
    \caption{Performance Comparison: Deployment Feasibility \& Benchmarking}
    \label{table:baseline_vs_ultra_tiny}
    \resizebox{\textwidth}{!}{
    \begin{tabular}{lcccccccccc} % Fixed: Changed from 12 to 11 columns to match data
        \toprule
        % Fixed: Added row counts and width rules {*} to all multirows
        \multirow{2}{*}{\textbf{Model Config}} & 
        \multicolumn{2}{c}{\textbf{Flash Memory (B)}} & 
        \multicolumn{3}{c}{\textbf{SRAM Usage (B)}} & 
        \multirow{2}{*}{\textbf{\#MACs}} & 
        \multirow{2}{*}{\textbf{\#BOPs}} & 
        \multirow{2}{*}{\textbf{Latency (s)}} & 
        \multirow{2}{*}{\textbf{Energy (J)}} & 
        \multirow{2}{*}{\textbf{Accuracy (\%)}} \\
        \cmidrule(lr){2-3} \cmidrule(lr){4-6}
        & \textbf{Weights} & \textbf{Code} & 
        \textbf{Global Data} & \textbf{Workspace} & \textbf{Stack} & 
        & & & & \\
        \midrule
        \multicolumn{11}{l}{\textit{\textbf{I. Target Deployment (Hardware: ATmega328P)}}} \\
        \textbf{DMC-Tiny (Ours)} & \textbf{16,384} & \textbf{10,578} & \textbf{1,588} & \textbf{1,296} & \textbf{78} & \textbf{131,040} & \textbf{1,619,428} & \textbf{2.80} & \textbf{0.308} & \textbf{98.17} \\
        \midrule
        \multicolumn{11}{l}{\textit{\textbf{II. Comparative Benchmark (Proxy Hardware: RP2040/Pico)$^{\dagger}$}}} \\
        Baseline (FP32) & 148,242 & \textbf{43,604} & 25,288 & 5,880 & \textbf{532} & 392,040 & \textbf{0} & 5.58 & 15.624 & \textbf{99.42} \\
        DMC-Ultra & \textbf{2,640} & 43,876 & 7,712 & 5,880 & 544 & 226,740 & 2,269,400 & 0.15 & 0.431 & 98.77 \\
        DMC-Tiny (Ref) & 16,384 & 43,916 & \textbf{3,128} & \textbf{1,296} & 552 & \textbf{131,040} & 1,619,428 & \textbf{0.09} & \textbf{0.257} & 98.17 \\
        \midrule
        \textit{Reduction (Base vs Ultra)} & \textit{56.2$\times$$^1$} & \textit{$\sim$1.0$\times$} & \textit{$\sim$1.0$\times$} & \textit{3.3$\times$} & \textit{-} & \textit{1.7$\times$} & \textit{N/A} & \textit{36$\times$} & \textit{36$\times$} & \textit{-0.6\%} \\
        \bottomrule
    \end{tabular}
    }
    \smallskip % Adds a tiny, clean space before the footnote text
    
    \footnotesize{$^1$Weights loaded from flash. \textbf{$^{\dagger}$Baseline and Ultra models exceed ATmega resources; metrics measured on Raspberry Pi Pico (RP2040) with software-emulated FPU for fair comparison. Code size on Pico is higher due to SDK overhead.}}
\end{table*}

% To validate the proposed DMC pipeline, we conducted a rigorous evaluation focusing 
Our computational experiments focus on deployment feasibility across the hardware spectrum. We utilize the LeNet-5 architecture on MNIST as a primary case study to analyze the interaction between compression algorithms and bare-metal constraints. \textcolor{black}{Though LeNet-5 is considered a trivial benchmark for 32-bit platforms, it represents a `worst-case' stress test for 8-bit bare-metal deployment. Its initial convolutional layers generate high activation volumes that exceed the 2KB SRAM limit of common MCUs like the ATmega328P by nearly 3×. We utilize this architecture to evaluate DMC’s ability to reconcile standard CNN structures with extreme hardware scarcity.} We aim to answer the following questions:
\begin{itemize}
    \item\textbf{Compression and Accuracy:} What is the minimum achievable model size without significant accuracy degradation?
    \item\textbf{Resource Feasibility:} Can the pipeline enable deployment on devices previously considered too constrained for deep networks (e.g., $<$2KB SRAM)? 
    \item\textbf{System Efficiency:} How does the compressed model perform against the baseline? How does the dependency-free code compare to standard frameworks (TFLite)?
\end{itemize}

\subsection{Metrics and Setup} 
To comprehensively evaluate the proposed pipeline, we track three categories of performance metrics, comparing all results against an uncompressed 32-bit floating-point baseline:

\begin{itemize}
    \item \textbf{Memory Footprint:} 
    We measure \textbf{Model Binary Size} (Flash usage for weights and instruction code) and \textbf{Peak SRAM Usage} (the static buffers, stack overhead, and \textbf{Activation Workspace} required during a forward pass). These metrics determine the feasibility of deployment on constrained devices such as the ATmega328P. 
    \item \textbf{Computational Complexity:} We quantify the algorithmic workload using Multiply–Accumulate operations (\textbf{\#MACs}) and Bit Operations (\textbf{\#BOPs}). Lower values indicate reduced and memory processing requirements. 
    \item \textbf{Deployment Performance:} We report \textbf{Inference Latency} and \textbf{Estimated Energy Usage}
    \footnote{Estimated Energy Usage per inference is computed as the product of measured inference latency and the device's power.}
per inference to validate real-time efficiency. Finally, we track \textbf{Top-1 Accuracy} to assess the trade-off between aggressive compression and model fidelity. 
    
\end{itemize}

\subsection{LeNet-5 on MNIST} 
We first evaluated the pipeline on LeNet-5 (Baseline: 99.42\% accuracy, 145KB size). Following a layer-sensitivity analysis
% (see \textcolor{orange}{(should this be left here?)Appendix} Figure~\ref{fig:lenet5_sparsity})
, we performed a neural architecture search to identify optimal pruning and quantization configurations for two distinct objectives:

\begin{enumerate}
    \item \textbf{DMC-Ultra (Max Compression):} Optimized for minimum weight storage to test the limits of the pipeline. 
    \item \textbf{DMC-Tiny (Feasibility):} Optimized to satisfy the strict 2KB SRAM constraint of the ATmega328P while maximizing accuracy. 
\end{enumerate}

\subsubsection{Compression Performance} 
As shown in Table~\ref{table:baseline_vs_ultra_tiny}, \textbf{DMC-Ultra} achieved a $55.8\times$ reduction in parameter size (2.66KB vs. 145KB) with a minimal accuracy drop of 0.65\% (98.77\%). However, Table~\ref{table:baseline_vs_ultra_tiny} reveals a major bottleneck for deployment. While DMC-Ultra fits best into the 32KB Flash of the ATmega328P, its peak activation memory (5.74 KB) violates the device's strict 2KB SRAM limit. This highlights that weight compression alone is insufficient for bare-metal deployment, activation memory is often a harder constraint.

To address this, the DMC-Tiny configuration utilizes aggressive early-layer pruning to reduce the peak activation workspace to 1.27 KB. This leaves ample headroom ($\approx700 $bytes) for the system stack and drivers, enabling, to the best of our knowledge, the first documented deployment of a standard CNN on an Arduino Uno with $98.17\%$ accuracy.

\subsubsection{Comparison with Prior Bare-Metal Implementations} 
Compared to prior bare-metal implementations on the ATmega328P, DMC offers 
a strong balance of accuracy, memory, and generality. While LogNNet \cite{Izotov_2021} fits within memory, it suffers from low accuracy (84\%) and extreme latency (7.11s). Conversely, Gural et al. \cite{pmlr-v97-gural19a} achieved high accuracy (99.11\%) with a significantly lower latency of 684ms. However, this performance is dependent on domain-specific optimizations that sacrifice generality. The reduced latency discrepancy stems from three critical design choices in \cite{pmlr-v97-gural19a}: the use of 16-bit accumulation (vs. DMC's numerically stable 32-bit), downsampling inputs to 14×14 (reducing MACs by $\approx30\%$ vs. our full-resolution 28×28), and fitting the model entirely in single-cycle SRAM. In contrast, DMC streams weights from Flash to support larger, more expressive models. While adopting these constraints would align our latency, we prioritize architectural universality over single-task specialization.
% (2.8s vs. 684ms)
% \subsubsection{Deployment on ATmega328P} 
% To validate bare-metal feasibility, we deployed \textbf{DMC-Tiny} on an Arduino Uno (ATmega328P). This configuration satisfied the board's strict hardware constraints (2KB SRAM, 32KB Flash) while retaining an accuracy of \textbf{98.xx\%}.

% Table~\ref{table:full_dmc_board_result}\footnote{Deployment of DMC-Ultra and the baseline were done on an ATmega2560 as the baseline cannot fit on the ATmega328P, DMC-Tiny was deployed on ATmega328P} compares our result against prior bare-metal implementations. While LogNNet \cite{Izotov_2021} fits within memory, it suffers from low accuracy (84\%) while incur massive latency (7.11s). Conversely, Gural et al. \cite{pmlr-v97-gural19a} reported high accuracy (99.11\%) by utilising direct convolution which does inplace convolution and downsamples the images to 14 by 14,  and recoreded a latency of 684ms. DMC achieves a latency of 2.6s with 28 by 28.
% the best trade-off, maintaining 98\%+ accuracy with a latency of just \textbf{22ms}, representing a 29× speedup over \cite{pmlr-v97-gural19a}.

\section{Conclusion}

This work presented Deep Microcompression (DMC), a framework for deploying neural networks on bare-metal microcontrollers without heavy runtime dependencies. By successfully integrating structured pruning, QAT, and a hardware-aware bit-packing scheme, DMC achieves significant compression ratios (up to $55.8\times$ for LeNet-5). Benchmarks against TFLite highlight DMC’s superior footprint and portability, enabling workload migration from 32-bit to energy-efficient 8-bit devices.
Future work will explicitly target Transformer-based architectures, furthering our goal to provide locally relevant, accessible, and high-impact AI solutions for low-resource communities worldwide.

\newpage
\bibliography{ms}
\bibliographystyle{icml2026}

% \newpage
\clearpage
\appendix
\section{Preliminaries and Related Work}
The field of model compression has a rich history, driven by the dual goals of reducing the footprint of the model and accelerating inference. This section provides a comprehensive overview of key compression techniques that serve as the foundation for the proposed DMC pipeline, critically examining their historical context and modern applications, particularly in the domain of TinyML.

\subsection{Pruning}
Pruning, the process of removing redundant connections or neurons from a neural network, dates back to early foundational methods. These include "Optimal Brain Damage" and "Optimal Brain Surgeon," which analyzed the second derivative of the loss function to identify and remove the least important weights \cite{NIPS1989_6c9882bb, Hassibi1993OptimalBS}. The modern resurgence of pruning was catalyzed by the \emph{Deep Compression} framework by Han et al. \cite{han2015deep} which showed that significant weight reduction could be achieved through an iterative pruning, retraining, and quantization process.

A key distinction in pruning is between unstructured and structured methods. Unstructured pruning arbitrarily removes individual weights, offering maximum selection flexibility and the highest possible compression ratios \cite{frankle2019lotterytickethypothesisfinding, zhu2017prunepruneexploringefficacy} 
% \textcolor{red}{.(this can be removed to reduce by one line)The best compression result is seen when unstructured pruning is used}
, although this often comes with more computation cost \cite{9302790, gale2019statesparsitydeepneural, 10.1145/3613424.3614303}. Structured pruning removes entire groups of parameters, such as channels, filters, or layers, resulting in a smaller but still dense model \cite{li2017pruningfiltersefficientconvnets, Luo_2017_ICCV, Liu_2017_ICCV, liu2018rethinking}. Methods like ThiNet \cite{Luo_2017_ICCV} and Network Slimming \cite{Liu_2017_ICCV} use channel-level scaling factors to identify and remove redundant channels. Channel pruning, in particular, offers a balance between compression and hardware efficiency. The resulting model is a subset of the original model with the same type but with reduced operations. The remaining model is a standard network architecture that can be deployed on any platform without specialized sparse matrix libraries. This inherent hardware-friendliness makes structured pruning a critical component of any compression pipeline targeting microcontrollers. Our approach adopts this approach to ensure a dense, highly efficient model after pruning.

\subsection{Quantization}
% \textcolor{red}{(not needed, everyone (should) know this)Quantization is the conversion of a value to a finite set of values, a process that reduces the number of bits required to represent a weight or activation.} 
Most neural networks are traditionally trained and deployed using floats, which require at least 32 bits to be stored in memory. This has been the standard due to the precision required for convergence and accuracy, though lower-precision formats like 16-bit and 8-bit floats are increasingly used \cite{s22031230_floating_point}. 

There are several methods for implementing quantization. Integer types have been used as a look-up for parameters, where parameters are stored as an index for a code book and retrieved during inference \cite{han2015deep}. This suffers from a similar computational problem to floats, as the operation performed is in floating point which we aim to avoid in favor of integer operations. Other approaches, such as \cite{li2022bitserialweightpoolscompression}, have proposed techniques such as group weight pool to optimize the lookup process and reduce the number of weight operations. 

To avoid the overhead associated with look-ups, dynamic and static quantization are employed, this approach requires one extra parameter (scale) for symmetric quantization and two (scale and zero point) for asymmetric quantization. We employ static quantization, which pre-computes scaling factors to enable pure integer arithmetic, avoiding the runtime overhead of dynamic quantization. Model size and performance after quantization depend on the bitwidth. The move to extremely low bitwidths, such as binary (1-bit) or ternary (2-bit) networks, represents the extreme end of quantization \cite{courbariaux2016binarizedneuralnetworkstraining, zhu2017trainedternaryquantization}. Furthermore, we adopt Quantization-Aware Training (QAT) \cite{jacob2018quantization} over Post-Training Quantization (PTQ) \cite{migacz20178} to mitigate accuracy degradation.

\subsection{Bit-Packing and Low-Level Optimizations}

While pruning and quantization reduce model parameter count, the final step of efficiently storing the low-bitwidth weights in memory is critically important for TinyML deployment. Prior work has explored sub-byte storage but often introduces computational overheads unsuitable for generic microcontrollers. The Deep Compression framework \cite{han2015deep} utilizes Huffman coding, a variable-length scheme that necessitates complex serial decoding, introducing non-deterministic latency. More recent hardware-centric approaches have proposed alternative layouts: \cite{li2022bitserialweightpoolscompression} introduces bit-serial weight pools, with a custom lookup process that iterates over input bits, complicating the standard convolution kernel. Similarly, \cite{cowan2018automatinggenerationlowprecision} utilizes bit-plane decomposition, effectively creating a new axis of addressing that increases memory access irregularity while also requiring to iterate over input bits for computation.

In contrast, our approach prioritizes architectural universality. We implement fixed-length bit-packing, mapping multiple low-precision weights (e.g., four 2-bit weights) directly into standard memory units (8-bit bytes or 32-bit words). Unlike bit-serial approaches that require specialized execution kernels, our structure allows for decompression via elementary bitwise shift and mask operations, ensuring that the compressed model can be executed on any standard microcontroller architecture using generic integer instructions and offering a balance between theoretical compression ratios and execution efficiency.

\begin{algorithm}[tb!]
    \caption{Runtime Bit-Unpacking}
    \label{algorithm:bit_unpack_method}
    \begin{algorithmic}[1]
        \STATE \textbf{Input:} Index $i$; packed tensor $T_P$; bitwidth $b$
        \STATE \textbf{Constants:} $n_b \gets \lfloor 8/b \rfloor$; \quad $M \gets (2^b) - 1$
        \STATE $b_{offset} \gets (i \bmod n_b) \times b$
        \STATE $B \gets T_P[i \div n_b]$
        \STATE $B \gets (B \gg b_{offset})\ \&\ M$\IF{is\_signed}
            \STATE $P \gets \textsc{SignExtend}(B,\, b)$
        \ELSE
            \STATE $P \gets B$
        \ENDIF
        \STATE \textbf{return} $P$
    \end{algorithmic}
\end{algorithm}

\begin{algorithm}[htb!]
    \caption{Compile-time Bit Packing}
    \label{algorithm:pack params into byte}
    \begin{algorithmic}[1]
        \STATE \textbf{Input:} Parameter list $P = (p_1, p_2, \dots, p_{n_b})$; bitwidth $b$
        \STATE \textbf{Require:} $\text{len}(P) \leq 8 \div b$
        \STATE $shift \gets b$; \quad $mask \gets (2^b) - 1$; \quad $B \gets 0$
        \FOR{each value $v$ in $\text{reverse}(P)$}
            \STATE $B \gets (B \ll shift)\ |\ (v\ \&\ mask)$
        \ENDFOR
        \STATE \textbf{return} $B$
    \end{algorithmic}
\end{algorithm}

\subsection{Impact of Toolchain Optimization} 
To determine the optimal build configuration for bare-metal deployment, we performed an ablation study on the DMC-Ultra model using avrgcc, compilers with varying optimization levels. The results are detailed in Table~\ref{table:compiler_ablation_study}.

\textbf{Code Size vs. Latency Trade-off:} As illustrated in Table~\ref{table:compiler_ablation_study}, optimization flags have a decisive impact on deployability. The unoptimized build (avr-gcc -O0) resulted in a binary size of 17.1KB, which can exceed the available flash memory of 8-bit devices. Enabling size optimization (-Os) reduced the instruction footprint by approximately 38.7\%, making it best viable configuration for the ATmega328P. In terms of computational efficiency (-O3) gives the best latency and less energy consumption per inference which is of large interest for real world deployment.

\textbf{Memory Overhead:} Crucially, the static RAM usage (Global variables) remains largely constant across optimization levels ($\approx6 KB$ which is dominated by the preallocated activation workspace 5.74 KB). However, stack usage varies significantly. The -Os flag minimized stack depth (by 23\%) by aggressively inlining bit-unpacking routines, ensuring that the DMC-Tiny model stays within the strict 2KB SRAM budget (Peak: Static+Stack $<$ 2048 B).

\subsection{Impact of Packing Operation} As detailed in Table~\ref{table:compiler_ablation_study}, we analyzed the runtime overhead of the bit-unpacking routine by comparing the 4-bit DMC-Ultra configuration (4W4A) against a byte-aligned 8-bit variant (8W8A). While 4-bit packing reduces weight storage by 56\% 
and Static RAM by 48\%, allowing deployment on more constrained devices, it introduces more than $20\times\#\texttt{MAC}$ additional Bit Operations (BOPs) for on the-fly decoding workload. Consequently, inference latency increases by 46\% 
and energy consumption rises proportionally. This identifies software-based unpacking as a major bottleneck, suggesting a need for more efficient packing and unpacking methods and specialized hardware support (e.g., barrel shifters) in future low-power MCUs.

\begin{table*}[h]
    \centering
    \caption{Ablation Study: Impact of Toolchain and Bit-Packing (Hardware: ATmega2560)}
    \label{table:compiler_ablation_study}
    \resizebox{\textwidth}{!}{
    \begin{tabular}{lcccccccccc} % Fixed: Changed from 14 columns to 11 to match data
        \toprule
        % Fixed: Added row count '2' and width requirement '*' to all multirows
        \multirow{2}{*}{\textbf{Configuration}} & \multicolumn{2}{c}{\textbf{Flash Memory (B)}} & \multicolumn{3}{c}{\textbf{SRAM Usage (B)}} & \multirow{2}{*}{\textbf{\#MACs}} & \multirow{2}{*}{\textbf{\#BOPs}} & \multirow{2}{*}{\textbf{Latency (s)}} & \multirow{2}{*}{\textbf{Energy (J)}} & \multirow{2}{*}{\textbf{Accuracy (\%)}} \\
        \cmidrule(lr){2-3} \cmidrule(lr){4-6}
        & \textbf{Weights} & \textbf{Code} & \textbf{Global Data} & \textbf{Workspace} & \textbf{Stack Peak} & & & & \\
        \midrule
        \multicolumn{11}{l}{\textit{\textbf{I. Compiler Optimization (Model: DMC-Ultra, Pack: 4-bit)}}} \\ % Fixed: Changed \multicolumn from 9 to 11
        avr-gcc -O0 (None) & 2,638 & 17,484 & \textbf{6,158} & 5,880 & 83 & 226,740 & 2,269,400 & 9.37 & 2.76 & 98.77 \\
        avr-gcc -O3 (Speed) & 2,638 & 13,566 & 6,168 & 5,880 & \textbf{64} & 226,740 & 2,269,400 & \textbf{4.77} & \textbf{1.41} & 98.77 \\
        avr-gcc -Os (Size) & 2,638 & \textbf{10,712} & 6,172 & 5,880 & 69 & 226,740 & 2,269,400 & 5.07 & 1.50 & 98.77 \\
        \midrule
        \multicolumn{11}{l}{\textit{\textbf{II. Bit-Packing Impact (Model: DMC-Ultra, Compiler: -Os)}}} \\ % Fixed: Changed \multicolumn from 9 to 11
        8-bit Packing (8W8A) & 5,090 & \textbf{10,634} & 6,172 & 5,880 & \textbf{66} & 226,740 & \textbf{0} & \textbf{3.97} & \textbf{1.17} & \textbf{98.95} \\
        \textbf{4-bit Packing (4W4A)} & \textbf{2,638} & 10,712 & \textbf{3,232} & \textbf{2,940} & 67 &  226,740 & 4,682,579 & 6.14 & 1.81 & 98.03 \\
        \bottomrule
    \end{tabular}
    }
\end{table*}

\begin{table*}[htb]
    \centering
    \caption{Benchmark: DMC vs. TFLite Micro (LeNet-5 on RP2040/Pico)}
    \label{table:tflite_comparison}
    \resizebox{1.25\columnwidth}{!}{
    \begin{tabular}{lcccccc}
        \toprule
        \textbf{Framework} & \textbf{Flash (KB)} & \textbf{RAM (KB)} & \textbf{Worksp. (KB)} & \textbf{Latency (s)} & \textbf{Energy (J)} & \textbf{Acc (\%)} \\
        \midrule
        \multicolumn{7}{l}{\textit{Float32 Baseline}} \\
        TFLite Micro (FP32) & 365.2 & 27.4 & 24.7 & 0.60 & 0.13 & 98.91 \\
        \textbf{DMC (FP32)} & \textbf{199.4} & \textbf{4.5} & 22.97 & \textbf{0.56} & 0.12 & 98.91 \\
        \midrule
        \multicolumn{7}{l}{\textit{Int8 Quantized}} \\
        TFLite Micro (Int8) & 254.9 & 10.7 & 8.4 & \textbf{0.055} & 0.012 & 98.88  \\
        \textbf{DMC (Int8)} & \textbf{86.4} & \textbf{7.7} & \textbf{5.7} & 0.226 & 0.05 & 98.89 \\
        \midrule
        \textit{DMC Advantage} & \textit{\textbf{3$\times$ Less}} & \textit{\textbf{1.4$\times$ Less}} & \textit{\textbf{1.5$\times$ Less}} & \textit{Slower} & \textit{-} & \textit{Equal} \\
        \bottomrule
    \end{tabular}
    }
\end{table*}

\subsection{Benchmarking Against TFLite for Microcontrollers}
To quantify the performance trade-offs of our dependency-free inference engine against specialized frameworks, we benchmarked DMC against Tensorflow Lite for Microcontrollers\cite{david2021tensorflowlitemicroembedded}. Experiments were conducted using a fp32 and quantized (W8A8) LeNet-5 model RP2040 (Cortex-M0+).

\textbf{Latency vs. Portability:} From Table~\ref{table:tflite_comparison} we observed that TFLite Micro achieves superior inference speed by leveraging \texttt{CMSIS-NN} optimized kernels. However, this hardware acceleration does not extend to floating-point operations, hence the similar latency. In contrast, DMC generates generic scalar C code. While this results in higher latency compared to platform-tuned libraries, it ensures execution on architectures lacking \texttt{CMSIS-NN} support.

\textbf{Binary Footprint Efficiency:} Despite the latency gap, DMC demonstrates a clear advantage in binary size. By eliminating the TFLite runtime interpreter and library overhead, DMC reduces the total binary size by $\approx3\times (168.5 KB)$. This saving is critical for ultra-constrained devices (e.g., ATmega328P) which possess only 32 KB of flash. 

These results reveal a fundamental trade-off between hardware-specific optimization and universal portability.
Explicitly trading ARM-specific speed for universality provides three critical advantages-Architectural Portability, Minimal Footprint, and Zero-Dependencies.

\end{document}